\documentclass[]{spie}
\usepackage{graphicx, caption, subcaption}
\usepackage[]{times}
\usepackage[]{url}
\usepackage[sort, numbers]{natbib}
\usepackage[nohyperlinks, printonlyused, withpage, smaller, nolist]{acronym}
\usepackage{hyperref}
\usepackage{gensymb}
\usepackage{rotating}
\usepackage[export]{adjustbox}

\title{ Where to drive? \linebreak
Free Space Detection with one Fisheye Camera
}

\author{Tobias Scheck, Adarsh Mallandur, Christian Wiede, Gangolf Hirtz
\skiplinehalf
Chemnitz University of Technology
}

\begin{document}
\maketitle

\begin{abstract}

The development in the field of autonomous driving goes hand in hand with ever new developments in the field of image processing and machine learning methods. In order to fully exploit the advantages of deep learning, it is necessary to have sufficient labeled training data available. This is especially not the case for omnidirectional fisheye cameras. As a solution, we propose in this paper to use synthetic training data based on Unity3D. A five-pass algorithm is used to create a virtual fisheye camera. This synthetic training data is evaluated for the application of free space detection for different deep learning network architectures. The results indicate that synthetic fisheye images can be used in deep learning context.

\end{abstract}


\begin{acronym}
\acro{SGD}{stochastic gradient descent}
\acro{CNN}{convolutional neural network}
\acro{IoU}{Intersection-over-Union}
\end{acronym}

\section{Introduction}
\label{sec:intro}
In recent years, the idea of autonomous driving has gained tremendous momentum.
All automotive companies and suppliers have set up large research departments to turn the dream of driver-less driving into reality.
The development is currently between levels 2 (Partial automation) and 3 (Conditional automation) of the level of automation.
These includes systems such as adaptive cruise control, parking assistance or driving on highway.
In order to reach levels 4 (High Automation) and 5 (Full automation), however, further challenges have to be mastered.
A fundamental challenge is the detection of free spaces in which the car can generally move.
These can be streets, side walks or meadows but not parking vehicles, lanterns, trees or people occupying the place.
Therefore, it is essential for the navigation to detect these free spaces. This is not sufficiently fulfilled by existing systems.
In addition, many external sensors are required.

In this paper, we propose a way to detect free spaces using a single omnidirectional camera with a fisheye lens.
Fisheye cameras can capture a wide field of view.
This is at the expense of strong tangential and radial disturbances in the image.
For a proper training with machine learning algorithms a huge amount of training data is necessary.
However, less or no labeled training data exists for fisheye pictures.
Especially the semantic segmentation requires a pixel by pixel labeling.
This cannot be done manually without an extreme amount of resources.
Instead, we suggest to use synthetic data for the training.
If these are as realistic as possible, these images can be used for training and create a model working on real world data.
In order to create these images urban traffic scenes by Unity3D are used.
This environment can realize different backgrounds and dynamic driving scenarios.
Furthermore, we propose a five-pass algorithm in combination with synthetic data to generate virtual fisheye images and a transfer learned \ac{CNN} to generate a semantic segmentation.

This work is structured as follows:
In Sect. 2, the related work of deep learning based segmentation and synthetic data acquisition is outlined and the research gap highlighted.
The synthetic data generation by means of Unity3D is described in Sect. 3.
This is followed by the presentation of the deep learning in Sect. 4.
In Sect. 5, we present our experimental results, which is accompanied by a discussion. Finally, we summarise our outcomes and outline future work.

\section{Related Work}
\label{sec:related}
Identifying drivable areas and detection of surrounding obstacles is a crucial task for autonomous driving.
A mixture of different sensors such as radar, lidar, camera, high precision GPS and prior map information is used to determine free space in current autonomous vehicles \cite{montemerlo2008junior,urmson2008autonomous}.
Research in autonomous driving started in Europe in 1986 with the PROMETHEUS project \cite{williams1988prometheus} involving more than 13 vehicle manufacturers and several universities from Europe.
VaMP driverless car was a notable result of this project which covered 95\,\% of about a distance of 1,600 km fully automatically \cite{dickmanns1997vehicles} and is considered as one of the first autonomous cars \cite{dickmanns2007dynamic}.
Projects such as these were usually computer vision systems for lateral and longitudinal vehicle guidance \cite{weber1995new, pomerleau1996rapidly}, lane departure warning \cite{franke1994daimler} or collision avoidance \cite{thorpe1988vision}.
The success of PROMETHEUS and other similar projects \cite{thorpe1988vision,morimoto1999ahs} influenced researchers to delve into more complex urban environments from the much simpler highway scenarios.
The term urban environment means that the focus is towards the problem of obstacle detection or free space detection rather than vehicle guidance.
Franke et al. \cite{franke1998autonomous} propose an "Intelligent Stop \& Go system", using depth-based obstacle detection and tracking from stereo images for urban environment.
In the 2000s, the problem of road detection was solved by using low-level features such as color, shape \cite{sotelo2004color} and texture \cite{lombardi2005switching} of the road.
Detection of road borders or road markings using laser \cite{sparbert2001lane}, radar \cite{ma2000simultaneous}, Hough transform \cite{yu1997lane} etc.
is also used as a technique to detect drivable space.
However, such algorithms fail under severe lighting variations and depend strongly on the structure of the roads.
Hence, more and more algorithms relied on geometric modeling using stereo images.
Badino et al. introduce a method for free space detection in complex scenarios using stochastic occupancy grids \cite{badino2007free}.
These grids are built using data from stereo cameras and are integrated over time with Kalman filters.
Occupancy grids are in a way segmentation between free and occupied regions.
Apart from segmentation of pixels, other representations such as Stixels \cite{badino2009stixel}, which groups objects with vertical surfaces, have also been used for the task of urban traffic scene understanding.

In the past decade, segmentation of road scenes using computer vision techniques has been considered as a standard approach to detect free spaces \cite{alvarez2012road}.
There are techniques ranging from Watershed Algorithms \cite{beucher1990road}, Dense Stereo Maps \cite{ladicky2012joint}, Structure from Motion (SfM) \cite{sturgess2009combining} to global methods such as Conditional Random Fields (CRF) \cite{passani2014crf} or Boosting Algorithms \cite{vitor2014probabilistic}.
Inspired from the success of \acp{CNN} in tasks such as classification, researchers have found ways to apply \acp{CNN} in the task of road segmentation \cite{alvarez2012road, brust2015convolutional, mohan2014deep}.
Alvarez et al. used \ac{CNN} based algorithm to recover 3D scene layout of a road image \cite{alvarez2012road}.
Hereby, the problems of overfitting and manual label generation are solved by using noisy labels generated from a classifier, which is trained on a general image dataset.
The authors in \cite{brust2015convolutional} introduce convolutional patch networks, which perform a pixel-wise labeling of input road scene images, by first applying patch segmentation or classification of image patches.
The deconvolution networks used in \cite{mohan2014deep} show a good performance on the KITTI benchmark.
However, this approach is computationally expensive and is not suited for real-time road segmentation \cite{oliveira2016efficient}.
The most notable work in the field of semantic segmentation are the fully convolutional networks (FCN) \cite{Shelhamer:2017:FCN:3069214.3069246}, which are realized by replacing fully connected layers of a \ac{CNN} by convolutional layers and then up-sampling the resulting feature map into a pixel-wise segmented image thus enabling end-to-end learning.
The absence of fully connected layers meant that the input can be of any size and that the spatial information is preserved.
Furthermore, the resulting architecture has a smaller number of trainable parameters, thus achieving better performance than patch networks.

Although state-of-the-art semantic segmentation networks achieve excellent accuracy, maximum information about the surrounding area is necessary for safe navigation.
Fisheye cameras provide a wider field of view than narrow-angle pin-hole cameras in complex urban traffic scenes and are becoming more popular in vehicles since they are cheap and easy to handle \cite{wang2014automatic, liu2008bird, haltakov2012scene}.
However, fisheye images are distorted due to strong perspective projections and are consequently unwarped for practical usage.
This unwarping process decreases the image quality, especially at the image boundaries.
Thus, \acp{CNN} trained on pin-hole camera images do not perform well on fisheye images.
Several research works have been carried out to develop algorithms that directly apply on distorted fisheye images \cite{fremont2016vision, deng2017cnn, deng2018restricted, saez2019real, qian2018pedestrian}.
The main problem of these algorithms is the lack of labeled fisheye urban scene datasets that have good quality and rich scenes.
Therefore, Qian et al. \cite{qian2018pedestrian} devise adversarial method of training \acp{CNN} using images from pin-hole camera images dataset to detect pedestrians in fisheye images.
An adversarial network generates fisheye pedestrian features, which are difficult for the detector to classify.
However adversarial networks are considered difficult to train because of various reasons such as unstable parameters, sensitivity to hyperparameters, mode collapses, vanishing gradients and overfitting \cite{arjovsky2017wasserstein}.
In \cite{deng2017cnn,deng2018restricted,saez2019real}, the authors synthetically generate fisheye images with labels from Cityscapes \cite{cordts2015cityscapes} and SYNTHIA \cite{ros2016synthia} datasets to train a \ac{CNN}.
The problem is that a transformation from a perspective to a fisheye image, loses image details in the centre during warping.
Our work will solve this problem by devising a method to generate realistic synthetic fisheye images along with the labels based on 5-pass rendering algorithm implemented in Unity3D.
It solves the problem of dataset generation as well as that of training a \ac{CNN} directly on distorted fisheye images by providing a way to generate fisheye datasets of different urban traffic scenes.

\section{Synthetic Data Creation}
In this section, we describe the generation of fisheye images along with their labels by a 5-pass rendering algorithm by using Unity3D.
Currently, different types of cameras such as perspective or omnidirectional cameras are used for autonomous driving.
Omnidirectional cameras provide a wider field of view (FOV) than perspective cameras thus eliminating the need for more cameras or mechanically rotatable cameras.
An ideal omnidirectional camera can capture light from all directions falling onto the focal point, thus covering a full sphere ($360\degree$ FOV).
There are various omnidirectional cameras ranging from catadioptric cameras with curved mirrors to dioptric cameras with purely dioptric fisheye lenses.
The name fisheye lens was coined by R. W. Wood to refer any lens capable of imaging the entire hemisphere in object space onto a finite circle in the focal plane \cite{kingslake1989history}.
Fisheye lenses are designed to cover the whole hemispherical field in front of the camera with a FOV of about $180\degree$.
Such fisheye lens cameras cannot be modeled by a linear projection due to very high distortions in the images caused by refractions in fisheye lenses.
The distortions in fisheye images where straight lines appear curved are known as radial symmetric distortions and are usually caused by the shape of the lens \cite{tardif2006self}.
Furthermore, these distortions are not uniformly distributed over all spatial areas \cite{fremont2016vision}.
There are several models to describe the behavior of fisheye lenses such as equidistant projection, stereographic projection, equisolid angle projection and orthogonal projection.
However, real lenses do not exactly follow the designed projection model \cite{kannala2006generic}.
Using such models, the authors in \cite{deng2017cnn, deng2018restricted, saez2019real, saez2018cnn} generated synthetic fisheye images from existing datasets with standard perspective projection images captured from a single camera.
Instead of transforming existing perspective projection images to fisheye projection images, our method directly synthesizes fisheye projection images using virtual cameras within Unity3D environment.

\begin{figure}[t]
\begin{center}
\includegraphics[width=\textwidth]{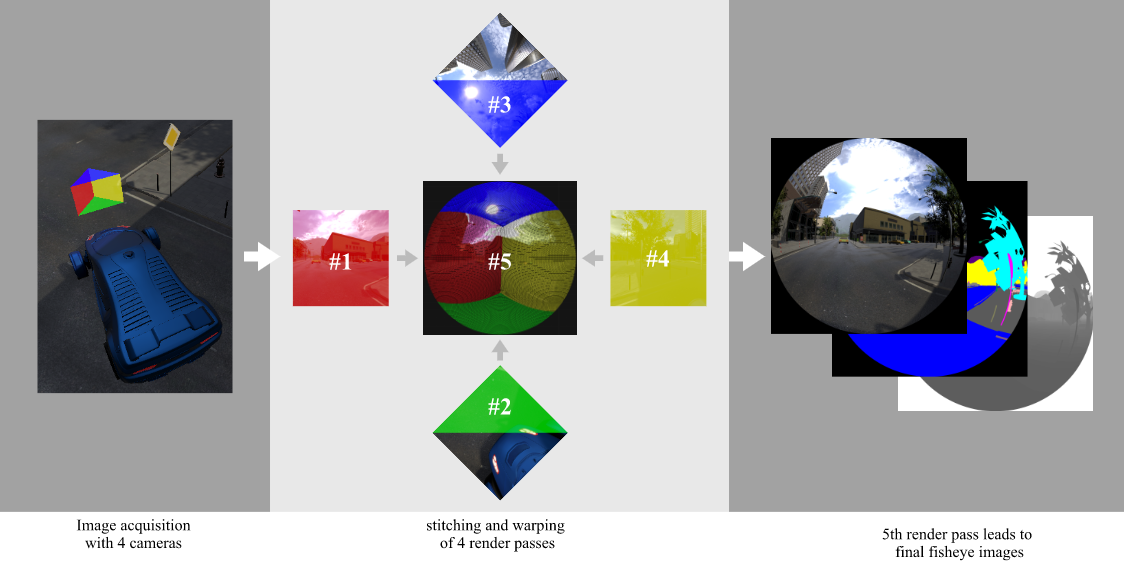}
\end{center}
\caption[example]
{ \label{fig:rendering}
Pipeline of synthetic data acquisition beginning with 4 cameras, each camera projection represents a face of a cube.
In the second step, each projection result is used as a texture for a precomputed mesh to form a fisheye image.
The previous steps are performed for each type of synthetic data (RGB image, segmentation mask and depth map) and exported as an image.
}
\end{figure}

In order to generate a dataset for training, Microsoft Airsim simulator \cite{shah2018airsim} designed for Unity3D is used.
Airsim provides a plugin with Python interface and a vehicle asset to be used in any Unity3D environment.
To help Artificial Intelligence (AI) research, Unity3D offers free Windridge City environment inside which Airsim asset vehicle can be imported.
The vehicle can be controlled either by keyboard or through the Airsim Python interface.
The generation of fisheye images is based on well-known technique of rendering to cubic maps, where a $360\degree$ view of the entire scene is rendered onto a cube.
In our case, we render to four faces of a cube by using each a perspective camera with a FOV of $90\degree$, as shown in \autoref{fig:rendering}.
The image captured by each camera is rendered into a texture, which is then used for the corresponding precomputed mesh to form a fisheye image.
The texture coordinates of these four meshes are designed in such a way that they capture the radial symmetric distortions and produce a fisheye projection.
Further information about the mesh generation can be obtained such as described in \cite{bourke2009idome}.
This is a computationally inexpensive method compared to a single pass rendering algorithm with a vertex shader that pre-distorts the geometry of the world to produce a fisheye projection \cite{bourke2009idome}.

In the five-pass rendering algorithm, four orthogonal cameras each with $90\degree$ FOV are created inside Unity3D and are placed above the virtual car.
The fifth camera is orthographic and is positioned to capture the fisheye projection formed by the combination of the four meshes.
During one frame update procedure of Unity3D, for each orthogonal camera, three additional orthogonal hidden cameras at the same position are created: one for the source image rendering, one for the label image rendering and one for the depth image rendering.
All these three hidden cameras render to the same texture as their parent cameras.
This approach is based on \cite{UnityTec68:online} and we modified it to use it with the five-pass rendering algorithm.
In order to render the label image, the layer id of each object is mapped to a color and is passed to the vertex shader for rendering.
The depth label is computed inside the vertex shader by calculating the normalized distance value of each pixel from the near plane and assigning it to that pixel intensity.
Therefore, for one image-saving cycle, three fisheye projections are obtained: source fisheye image, label fisheye image and depth fisheye image.
To preserve the floating-point values in the depth image, the .exr file format is used.

Each object inside Unity3D is assigned with a layer.
Thus, the total number of classes in the training set is equal to the number of layers in the Unity3D environment.
Unity3D supports a maximum of 32 layers.
These layer IDs along with their names and mapped colors are exported as JSON file in order to retain the mapping information.
The files were saved at a rate of one frame per second in order to reduce redundant images and the CPU load.
In order to train a classifier, 12028 images were generated by driving the virtual car around the Windridge city.
We call the generated dataset \textit{OmniCity} and the usage of a 80/20 training/validation split ratio results in 9623 training and 2405 validation images.
Thus, the proposed method provides a flexible way to generate a dataset in a urban driving environment.

\section{Deep Learning}
\label{sec:deeplearning}
This section introduces the \ac{CNN} free space detection architectures and the hyperparameters used for trainings.
The selected learning strategies and data augmentation methods during the trainings are described as well.

\begin{figure}[t]
\begin{center}
\includegraphics[width=\textwidth]{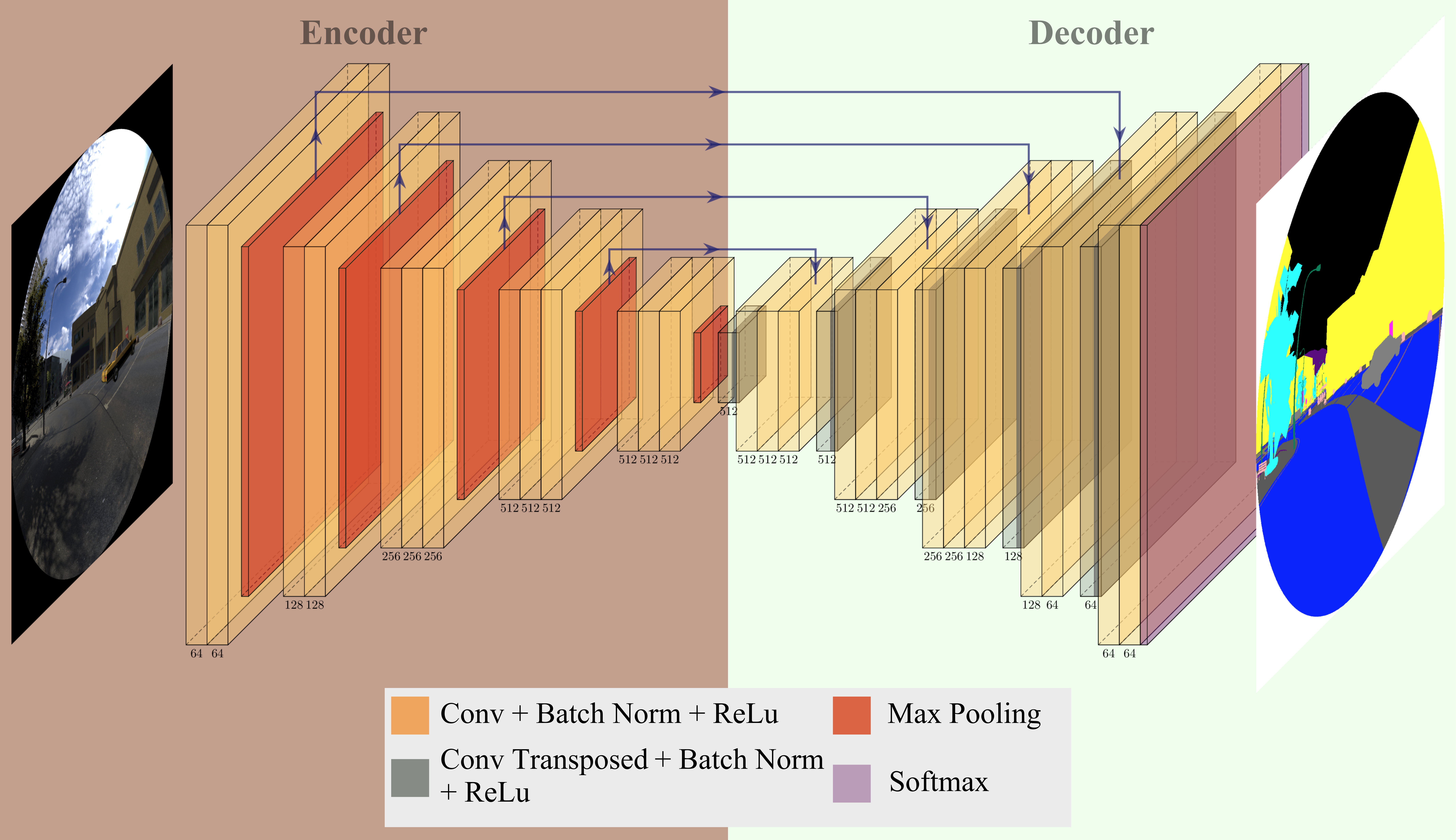}
\end{center}
\caption[example]
{ \label{fig:encoder-decoder}
This overview visualizes the used encoder-decoder structure for free space detection.
While the encoder produces sparse feature maps using max pooling, the decoder uses a transposed convolution for a learnable upsampling.
Visualization based on \cite{haris_iqbal_2018_2526396}
}
\end{figure}

The detection of free spaces in a scene requires a general image understanding on pixel level.
In our approach we are using \acp{CNN} designed for semantic segmentation tasks to train and classify pixel-wise.
As shown in \autoref{fig:encoder-decoder}, the first model is based on the implementation of SegNet \cite{badrinarayanan2017segnet}, which is an encoder-decoder architecture.
This architecture gradually reduces the spatial dimension in the encoder part, while the decoder part behaves in the opposite direction.
In concrete terms, this means as the \ac{CNN} progresses, object details and spatial dimensions are restored.

DeepLabV3 was chosen as additional \ac{CNN} architecture \cite{DBLP:journals/corr/ChenPSA17} with ResNet101 v2 \cite{he2016identity} as feature extractor.
Pooling layers, used in encoder-decoder architectures, are designed in such a way that they increase the field of view and aggregate necessary information but at the same time discards the local context.
However, this information is fundamental to semantic segmentation and should not be dispensed with.
To address this problem, the implementation of DeepLabV3 uses atrous convolutions.
They are also known as dilated convolutions and allow an enlargement of the field of view without reducing the spatial dimension and this at the same time efficiently \cite{yu2015multi,chen2017deeplab}.

Both architectures use RGB images as input with a resolution of $512\times512$ pixel.
By using ResNet101 v2 as feature extractor, weights, pre-trained on ILSVRC2012-CLS image classification dataset \cite{ILSVRC15}, are used for DeepLabV3.
The encoder-decoder weights are initialized using the technique described in Glorot et al. \cite{glorot2010understanding}.
To train all architectures we use \ac{SGD} with a momentum of 0.9.
An initial learning rate of 0.0001 for DeepLabV3 and 0.001 for the encoder-decoder architecture is selected.
A training process can benefit from a reduction in the learning rate during training progress.
To address this property, we use cosine decay \cite{loshchilov2016sgdr} as our learning rate strategy.
We train each architecture for 50 epochs with a mini-batch size of 4.
In the training process, we shuffle the dataset after each epoch to ensure that each training example is used only once during an epoch.
As objective function, cross-entropy loss is used for training which is summed over all pixel values in a mini-batch.
The last layer for each architecture calculates a per-pixel propability using softmax.
In this work, the \acp{CNN} distinguish only between free space and background.

In order to avoid over-fitting and ensure better generalization, data augmentation methods are applied during training \cite{takahashi2018data,krizhevsky2012imagenet}.
All methods depend on a random factor to decide a method is applied or not.
We have selected horizontal flipping, brightness changing with a max\_delta of 0.5 and random Gaussian noise with a mean of 0.0 and a standard deviation of 8.0.
For a training with the generated synthetic data we also change the hue of a RGB image and the saturation.

\section{Results and Discussion}
\label{sec:results}

In this section, we outline our experiments and discuss the corresponding result.
In the first experiment, we use the training set of the Cityscapes dataset \cite{cordts2015cityscapes}, containing 2975 images, to train both \ac{CNN} architectures on perspective images of urban street scenes.
Cityscapes contains segmentation masks for 30 classes.
However, for free space detection we map the classes: road, sidewalk, parking, rail track and terrain to the class \textit{free space} and everything else to the class \textit{background}.
The second experiment aims to simulate a fisheye distortion by mapping perspective images into fisheye images.
Accordingly all images of Cityscapes are distorted with a focal length of 159 based on the approach described in Deng et al. \cite{deng2017cnn}, named Fisheye Cityscapes.
In the third experiment, we use our approach for synthetic fisheye data generation to train all \acp{CNN} on OmniCity.

After training the architectures, we validate the performance using a sequence of real fisheye images.
For this purpose, a subset of the fisheye data set DriveA sequence \cite{eichenseer2016data} was manually labeled, in a coarse manner, for our free space scenario\footnote{https://gitlab.com/tschec/annotations}.
In order to evaluate the 150 labeled images of this subset, the \ac{IoU} was used.
The \ac{IoU} is described as the quotient of the area of overlap of the ground truth and detected mask and area of union of both masks.
During the evaluation all pixels beyond the fisheye FOV are labeled as void and ignored.

\begin{table}[h]
    \centering
    \caption{
    \label{tbl:results}
    IoU (\%) for free space detection after training using Cityscapes, Fisheye Cityscapes and our proposed synthetic dataset OmniCity.}
\begin{tabular}{|p{4cm}|c|c|}
\hline
                            & \multicolumn{2}{c|}{\ac{CNN}} \\ \hline
\multicolumn{1}{|c|}{Dateset} & Encoder-Decoder    & Deeplab v3   \\ \hline
Cityscapes                    &        93.1        &     84.6     \\ \hline
Fisheye Cityscapes            &        90.7        &     72.5     \\ \hline
OmniCity                      &        86.7        &     78.3     \\ \hline
\end{tabular}
\end{table}

As shown in \autoref{tbl:results}, the results for the \ac{CNN} architectures trained on Cityscapes are those that achieve the highest IoU.
To the contrary, the results for Fisheye Cityscapes decrease by 14.3\,\% for Deeplab v3 and by 2.58\,\% for the Encoder-Decoder structure.
It seems that the distortions caused by a fisheye lense do not affect the free space detection results.
This fact is supported by the results achieved by a training with OmniCity.
The generated synthetic data, following the equiangular projection, achieves comparable results to Fisheye Cityscapes.
The \ac{IoU} decreases, compared to Cityscapes, by 7.45\,\% using Deeplab v3 and about 6.87\,\% with the Encoder-Decoder.
However, it should be noted that, during training with OmniCity, none of the CNN architectures have ever seen real data.

An explanation for the poorer performance using fisheye distorted images can be the fact that the class free space is mainly not affected by those.
Unlike objects, it is difficult to recognize a pattern in free space.
Roads do not follow clear lines or terrain like meadows and fields rarely have flat surfaces, but these properties are most affected by distortion.
It can be assumed that for more complex classes such as cars, people or buildings, the distortion has a greater impact than that studied in our scenario.

\setlength{\fboxsep}{0pt}
\begin{figure*}[ht!]
\captionsetup[subfigure]{labelformat=empty}
\centering
    \begin{adjustbox}{minipage=\linewidth,scale=0.88}
    \begin{sideways}
        \hskip 0.05\textwidth
        \small \(Deeplab\ v3\)
    \end{sideways}
    \begin{subfigure}[b]{.24\linewidth}
        \caption{\(GT\)}
        \fbox{\includegraphics[width=\textwidth]{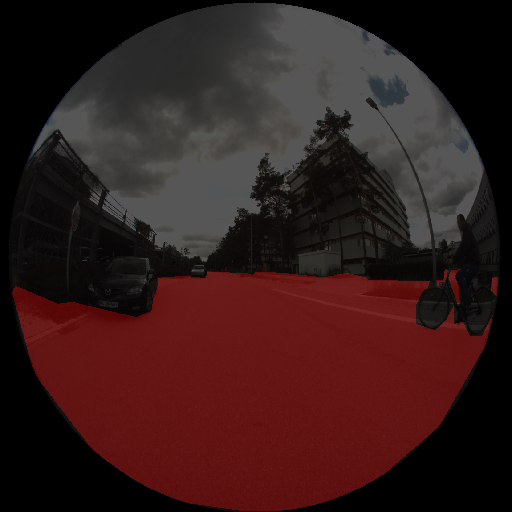}}
    \end{subfigure}
    \begin{subfigure}[b]{.24\linewidth}
        \caption{\(Cityscapes\)}
        \fbox{\includegraphics[width=\textwidth]{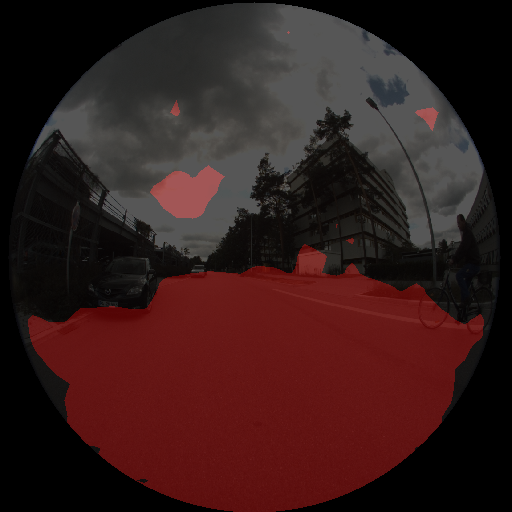}}
    \end{subfigure}
    \begin{subfigure}[b]{.24\linewidth}
        \caption{\(Fisheye\ Cityscapes\)}
        \fbox{\includegraphics[width=\textwidth]{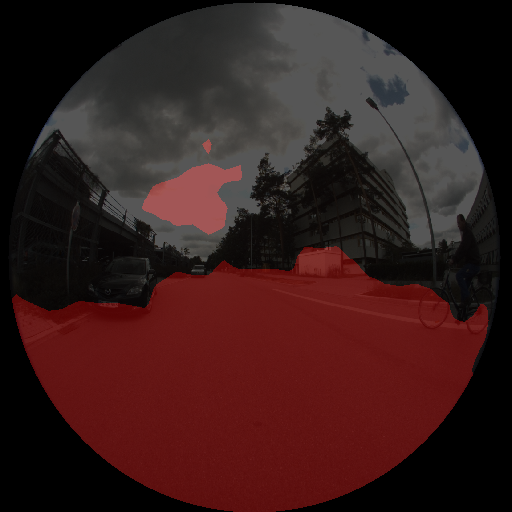}}
    \end{subfigure}
    \begin{subfigure}[b]{.24\linewidth}
        \caption{\(OmniCity\)}
        \fbox{\includegraphics[width=\textwidth]{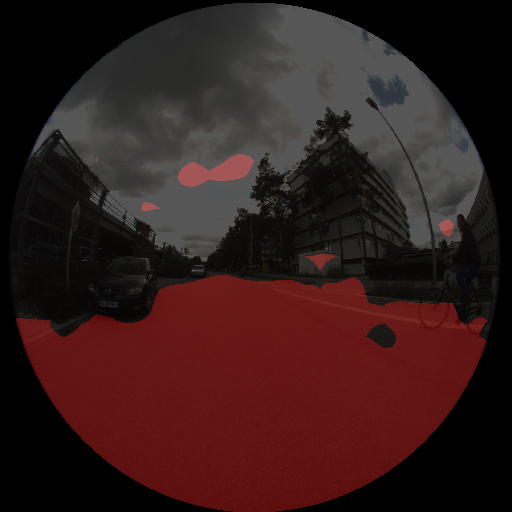}}
    \end{subfigure}

    \begin{sideways}
        \hskip 0.04\textwidth
        \small \(Encoder-Decoder\)
    \end{sideways}
    \begin{subfigure}[b]{.24\linewidth}
        \fbox{\includegraphics[width=\textwidth]{img/results/gt/DriveA_0006.png}}
    \end{subfigure}
    \begin{subfigure}[b]{.24\linewidth}
        \fbox{\includegraphics[width=\textwidth]{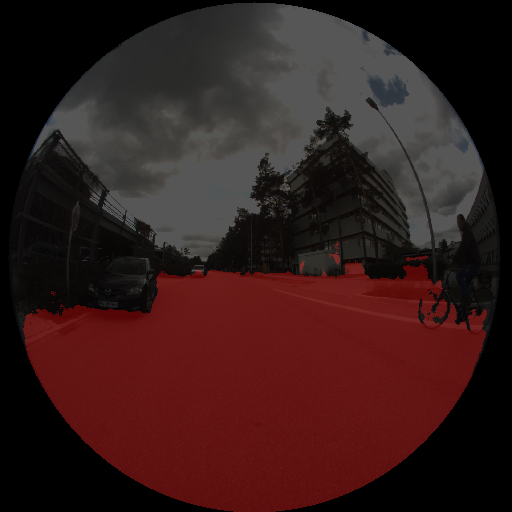}}
    \end{subfigure}
    \begin{subfigure}[b]{.24\linewidth}
        \fbox{\includegraphics[width=\textwidth]{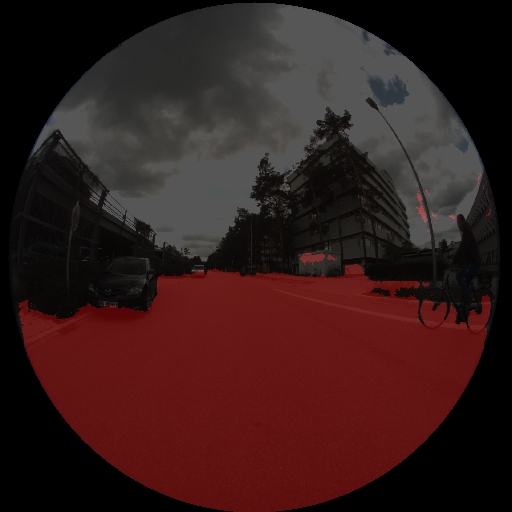}}
    \end{subfigure}
    \begin{subfigure}[b]{.24\linewidth}
        \fbox{\includegraphics[width=\textwidth]{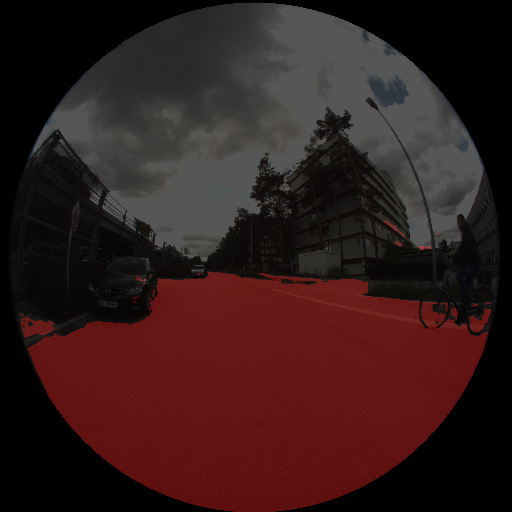}}
    \end{subfigure}

    \hskip 0.02\textwidth

    \begin{sideways}
        \hskip 0.05\textwidth
        \small \(Deeplab\ v3\)
    \end{sideways}
    \begin{subfigure}[b]{.24\linewidth}
        \fbox{\includegraphics[width=\textwidth]{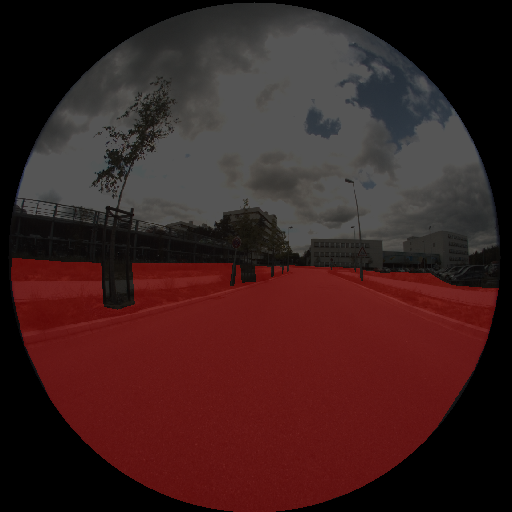}}
    \end{subfigure}
    \begin{subfigure}[b]{.24\linewidth}
        \fbox{\includegraphics[width=\textwidth]{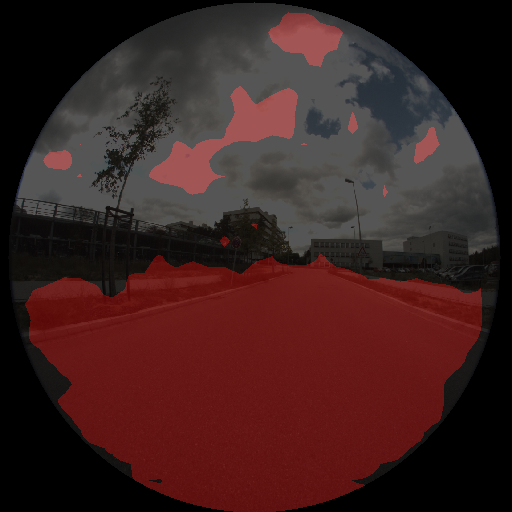}}
    \end{subfigure}
    \begin{subfigure}[b]{.24\linewidth}
        \fbox{\includegraphics[width=\textwidth]{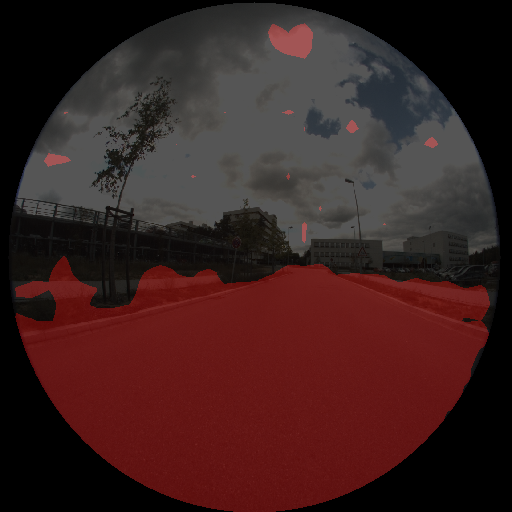}}
    \end{subfigure}
    \begin{subfigure}[b]{.24\linewidth}
        \fbox{\includegraphics[width=\textwidth]{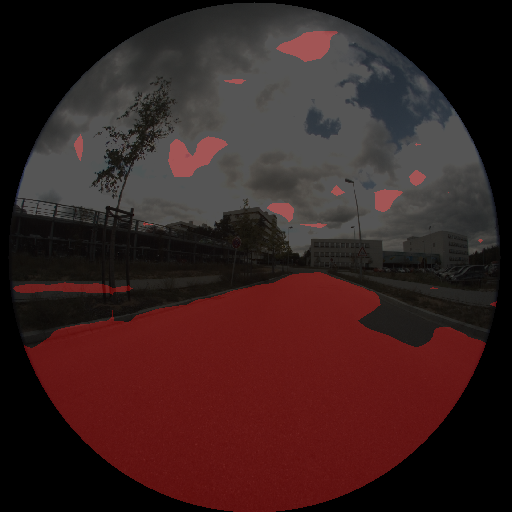}}
    \end{subfigure}

    \begin{sideways}
        \hskip 0.04\textwidth
        \small \(Encoder-Decoder\)
    \end{sideways}
    \begin{subfigure}[b]{.24\linewidth}
        \fbox{\includegraphics[width=\textwidth]{img/results/gt/DriveA_0165.png}}
    \end{subfigure}
    \begin{subfigure}[b]{.24\linewidth}
        \fbox{\includegraphics[width=\textwidth]{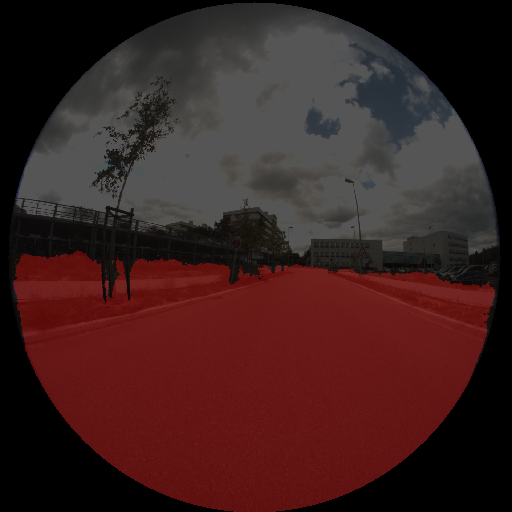}}
    \end{subfigure}
    \begin{subfigure}[b]{.24\linewidth}
        \fbox{\includegraphics[width=\textwidth]{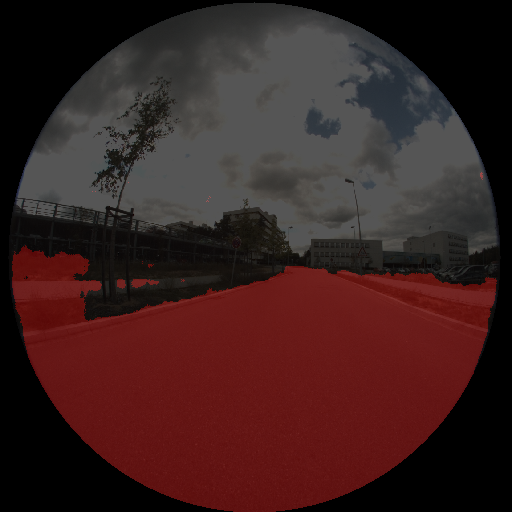}}
    \end{subfigure}
    \begin{subfigure}[b]{.24\linewidth}
        \fbox{\includegraphics[width=\textwidth]{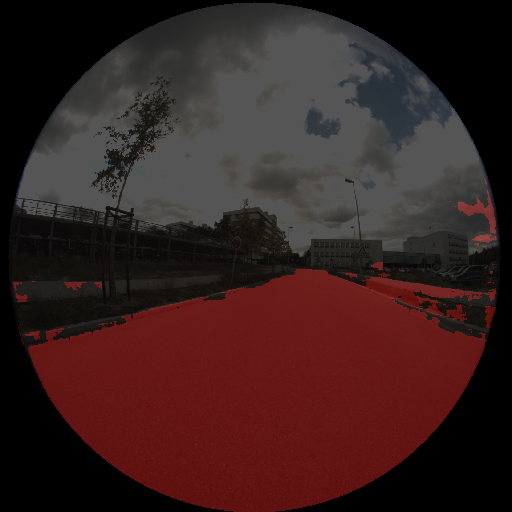}}
    \end{subfigure}
    \end{adjustbox}
    \caption{Qualitative results of the evaluated \ac{CNN} architectures, trained on different datasets. \label{fig:results}}
\end{figure*}

In \autoref{fig:results}, qualitative free space segmentation results for the experiments are shown.
In general, Deeplab v3 is found to be slightly worse than the Encoder-Decoder architecture.
However, this also reflects the results from \autoref{tbl:results}.
In the first selected scenario (row 1 and 2), a recognizable gap of quality between the results of all Encoder-Decoder architectures seems minimal.
The difference in segmentation quality only becomes apparent in the last scenario (row 3 and 4).
While the CNNs trained with Cityscapes and Fisheye Cityscapes are able to segment roads, footpaths and even terrain between them, the model trained with OmniCity is not detecting these accurately.
This is particularly evident in the grassy areas and footpaths.
One possible cause may be a lack of training examples with such constellations.

However, we could demonstrate that it is possible to learn deep learning architectures based on synthetic fisheye data and are able to apply that model to real world data.
Nevertheless, further work as to be carried out in this field.
\section{Conclusion}
In this work, we proposed an approach for free space detection using a single fisheye camera to benefit from its wider FOV.
Due the lack of omnidirectional training data, we use a data augmentation method, introduced in \cite{deng2017cnn}, to simulate a fisheye distortion on perspective images.
Additionally, we described a method to generate synthetic omnidirectional images, segmentation masks and depth maps, using Unity3D.
A new omnidirectional dataset, called OmniCity, with urban synthetic city scenes is introduced.
Furthermore, an Encoder-Decoder and the Deeplab v3 \ac{CNN} architecture in the context of free space detection was evaluated, which is commonly used for semantic segmentations tasks.

It can been shown that for free space detection on omnidirectional images, the use of fisheye distorted images for training is not mandatory, but can be advantageous in case no real world data is available.
However, during our evaluation we have shown that \acp{CNN} trained only on synthetic data can achieve comparable results.
In future research, the evaluation of more complex semantic segmentation scenarios should be investigated to address the usage of omnidirectional images.
For this purpose, we will improve our OmniCity dataset to include a higher variation in urban scenes, different light and weather conditions, different vegetation and more classes like people and additional vehicles.

\renewcommand{\refname}{REFERENCES}

\bibliographystyle{spiebib}
\bibliography{paper}

\begin{thebibliography}{10}

\bibitem{montemerlo2008junior}
Montemerlo, M., Becker, J., Bhat, S., Dahlkamp, H., Dolgov, D., Ettinger, S.,
  Haehnel, D., Hilden, T., Hoffmann, G., Huhnke, B., et~al., ``Junior: The
  stanford entry in the urban challenge,'' {\em Journal of field Robotics}~{\bf
  25}(9),  569--597 (2008).

\bibitem{urmson2008autonomous}
Urmson, C., Anhalt, J., Bagnell, D., Baker, C., Bittner, R., Clark, M., Dolan,
  J., Duggins, D., Galatali, T., Geyer, C., et~al., ``Autonomous driving in
  urban environments: Boss and the urban challenge,'' {\em Journal of Field
  Robotics}~{\bf 25}(8),  425--466 (2008).

\bibitem{williams1988prometheus}
Williams, M., ``Prometheus-the european research programme for optimising the
  road transport system in europe,'' in [{\em IEE Colloquium on Driver
  Information}{\nolinebreak\hspace{0.1em}]},   1--1, IET (1988).

\bibitem{dickmanns1997vehicles}
Dickmanns, E.~D. et~al., ``Vehicles capable of dynamic vision,'' in [{\em
  IJCAI}{\nolinebreak\hspace{0.1em}]},   {\bf 97},  1577--1592 (1997).

\bibitem{dickmanns2007dynamic}
Dickmanns, E.~D.,  [{\em Dynamic vision for perception and control of
  motion}{\nolinebreak\hspace{0.1em}]}, Springer Science \& Business Media
  (2007).

\bibitem{weber1995new}
Weber, J., Koller, D., Luong, Q.-T., and Malik, J., ``New results in
  stereo-based automatic vehicle guidance,'' in [{\em Proceedings of the
  Intelligent Vehicles' 95. Symposium}{\nolinebreak\hspace{0.1em}]},
  530--535, IEEE (1995).

\bibitem{pomerleau1996rapidly}
Pomerleau, D. and Jochem, T., ``Rapidly adapting machine vision for automated
  vehicle steering,'' {\em IEEE expert}~{\bf 11}(2),  19--27 (1996).

\bibitem{franke1994daimler}
Franke, U., Mehring, S., Suissa, A., and Hahn, S., ``The daimler-benz steering
  assistant: a spin-off from autonomous driving,'' in [{\em Proceedings of the
  Intelligent Vehicles' 94 Symposium}{\nolinebreak\hspace{0.1em}]},   120--124,
  IEEE (1994).

\bibitem{thorpe1988vision}
Thorpe, C., Hebert, M.~H., Kanade, T., and Shafer, S.~A., ``Vision and
  navigation for the carnegie-mellon navlab,'' {\em IEEE Transactions on
  Pattern Analysis and Machine Intelligence}~{\bf 10}(3),  362--373 (1988).

\bibitem{morimoto1999ahs}
Morimoto, H., Koizumi, M., Inoue, H., and Nitadori, K., ``Ahs road-to-vehicle
  communication system,'' in [{\em Proceedings 199 IEEE/IEEJ/JSAI International
  Conference on Intelligent Transportation Systems (Cat. No.
  99TH8383)}{\nolinebreak\hspace{0.1em}]},   327--334, IEEE (1999).

\bibitem{franke1998autonomous}
Franke, U., Gavrila, D., G{\"o}rzig, S., Lindner, F., Paetzold, F., and
  W{\"o}hler, C., ``Autonomous driving goes downtown,'' {\em IEEE Intelligent
  systems} (6),  40--48 (1998).

\bibitem{sotelo2004color}
Sotelo, M.~A., Rodriguez, F.~J., Magdalena, L., Bergasa, L.~M., and Boquete,
  L., ``A color vision-based lane tracking system for autonomous driving on
  unmarked roads,'' {\em Autonomous Robots}~{\bf 16}(1),  95--116 (2004).

\bibitem{lombardi2005switching}
Lombardi, P., Zanin, M., and Messelodi, S., ``Switching models for vision-based
  on-board road detection,'' in [{\em IEEE Conference on Intelligent
  Transportation Systems}{\nolinebreak\hspace{0.1em}]},   67--72 (2005).

\bibitem{sparbert2001lane}
Sparbert, J., Dietmayer, K., and Streller, D., ``Lane detection and street type
  classification using laser range images,'' in [{\em ITSC 2001. 2001 IEEE
  Intelligent Transportation Systems. Proceedings (Cat. No.
  01TH8585)}{\nolinebreak\hspace{0.1em}]},   454--459, IEEE (2001).

\bibitem{ma2000simultaneous}
Ma, B., Lakshmanan, S., and Hero, A.~O., ``Simultaneous detection of lane and
  pavement boundaries using model-based multisensor fusion,'' {\em IEEE
  Transactions on Intelligent Transportation Systems}~{\bf 1}(3),  135--147
  (2000).

\bibitem{yu1997lane}
Yu, B. and Jain, A.~K., ``Lane boundary detection using a multiresolution hough
  transform,'' in [{\em Proceedings of International Conference on Image
  Processing}{\nolinebreak\hspace{0.1em}]},   {\bf 2},  748--751, IEEE (1997).

\bibitem{badino2007free}
Badino, H., Franke, U., and Mester, R., ``Free space computation using
  stochastic occupancy grids and dynamic programming,'' in [{\em Workshop on
  Dynamical Vision, ICCV, Rio de Janeiro, Brazil}{\nolinebreak\hspace{0.1em}]},
    {\bf 20}, Citeseer (2007).

\bibitem{badino2009stixel}
Badino, H., Franke, U., and Pfeiffer, D., ``The stixel world-a compact medium
  level representation of the 3d-world,'' in [{\em Joint Pattern Recognition
  Symposium}{\nolinebreak\hspace{0.1em}]},   51--60, Springer (2009).

\bibitem{alvarez2012road}
Alvarez, J.~M., Gevers, T., LeCun, Y., and Lopez, A.~M., ``Road scene
  segmentation from a single image,'' in [{\em European Conference on Computer
  Vision}{\nolinebreak\hspace{0.1em}]},   376--389, Springer (2012).

\bibitem{beucher1990road}
Beucher, S., Bilodeau, M., and Yu, X., ``Road segmentation by watershed
  algorithms,'' in [{\em PROMETHEUS Workshop, Sophia Antipolis,
  France}{\nolinebreak\hspace{0.1em}]},  (1990).

\bibitem{ladicky2012joint}
Ladick{\`y}, L., Sturgess, P., Russell, C., Sengupta, S., Bastanlar, Y.,
  Clocksin, W., and Torr, P.~H., ``Joint optimization for object class
  segmentation and dense stereo reconstruction,'' {\em International Journal of
  Computer Vision}~{\bf 100}(2),  122--133 (2012).

\bibitem{sturgess2009combining}
Sturgess, P., Alahari, K., Ladicky, L., and Torr, P.~H., ``Combining appearance
  and structure from motion features for road scene understanding,'' in [{\em
  BMVC-British Machine Vision Conference}{\nolinebreak\hspace{0.1em}]},  BMVA
  (2009).

\bibitem{passani2014crf}
Passani, M., Yebes, J.~J., and Bergasa, L.~M., ``Crf-based semantic labeling in
  miniaturized road scenes,'' in [{\em 17th International IEEE Conference on
  Intelligent Transportation Systems (ITSC)}{\nolinebreak\hspace{0.1em}]},
  1902--1903, IEEE (2014).

\bibitem{vitor2014probabilistic}
Vitor, G.~B., Victorino, A.~C., and Ferreira, J.~V., ``A probabilistic
  distribution approach for the classification of urban roads in complex
  environments,'' in [{\em IEEE Workshop on International Conference on
  Robotics and Automation}{\nolinebreak\hspace{0.1em}]},  (2014).

\bibitem{brust2015convolutional}
Brust, C.-A., Sickert, S., Simon, M., Rodner, E., and Denzler, J.,
  ``Convolutional patch networks with spatial prior for road detection and
  urban scene understanding,'' {\em arXiv preprint arXiv:1502.06344}  (2015).

\bibitem{mohan2014deep}
Mohan, R., ``Deep deconvolutional networks for scene parsing,'' {\em arXiv
  preprint arXiv:1411.4101}  (2014).

\bibitem{oliveira2016efficient}
Oliveira, G.~L., Burgard, W., and Brox, T., ``Efficient deep models for
  monocular road segmentation,'' in [{\em 2016 IEEE/RSJ International
  Conference on Intelligent Robots and Systems
  (IROS)}{\nolinebreak\hspace{0.1em}]},   4885--4891, IEEE (2016).

\bibitem{Shelhamer:2017:FCN:3069214.3069246}
Shelhamer, E., Long, J., and Darrell, T., ``Fully convolutional networks for
  semantic segmentation,'' {\em IEEE Trans. Pattern Anal. Mach. Intell.}~{\bf
  39},  640--651 (Apr. 2017).

\bibitem{wang2014automatic}
Wang, C., Zhang, H., Yang, M., Wang, X., Ye, L., and Guo, C., ``Automatic
  parking based on a bird's eye view vision system,'' {\em Advances in
  Mechanical Engineering}~{\bf 6},  847406 (2014).

\bibitem{liu2008bird}
Liu, Y.-C., Lin, K.-Y., and Chen, Y.-S., ``Bird’s-eye view vision system for
  vehicle surrounding monitoring,'' in [{\em International Workshop on Robot
  Vision}{\nolinebreak\hspace{0.1em}]},   207--218, Springer (2008).

\bibitem{haltakov2012scene}
Haltakov, V., Belzner, H., and Ilic, S., ``Scene understanding from a moving
  camera for object detection and free space estimation,'' in [{\em 2012 IEEE
  Intelligent Vehicles Symposium}{\nolinebreak\hspace{0.1em}]},   105--110,
  IEEE (2012).

\bibitem{fremont2016vision}
Fremont, V., Bui, M., Boukerroui, D., and Letort, P., ``Vision-based people
  detection system for heavy machine applications,'' {\em Sensors}~{\bf 16}(1),
   128 (2016).

\bibitem{deng2017cnn}
Deng, L., Yang, M., Qian, Y., Wang, C., and Wang, B., ``Cnn based semantic
  segmentation for urban traffic scenes using fisheye camera,'' in [{\em 2017
  IEEE Intelligent Vehicles Symposium (IV)}{\nolinebreak\hspace{0.1em}]},
  231--236, IEEE (2017).

\bibitem{deng2018restricted}
Deng, L., Yang, M., Li, H., Li, T., Hu, B., and Wang, C., ``Restricted
  deformable convolution based road scene semantic segmentation using surround
  view cameras,'' {\em arXiv preprint arXiv:1801.00708}  (2018).

\bibitem{saez2019real}
S{\'a}ez, {\'A}., Bergasa, L.~M., L{\'o}pez-Guill{\'e}n, E., Romera, E.,
  Tradacete, M., G{\'o}mez-Hu{\'e}lamo, C., and del Egido, J., ``Real-time
  semantic segmentation for fisheye urban driving images based on erfnet,''
  {\em Sensors}~{\bf 19}(3),  503 (2019).

\bibitem{qian2018pedestrian}
Qian, Y., Yang, M., Wang, C., and Wang, B., ``Pedestrian feature generation in
  fish-eye images via adversary,'' in [{\em 2018 IEEE International Conference
  on Robotics and Automation (ICRA)}{\nolinebreak\hspace{0.1em}]},
  2007--2012, IEEE (2018).

\bibitem{arjovsky2017wasserstein}
Arjovsky, M., Chintala, S., and Bottou, L., ``Wasserstein generative
  adversarial networks,'' in [{\em International Conference on Machine
  Learning}{\nolinebreak\hspace{0.1em}]},   214--223 (2017).

\bibitem{cordts2015cityscapes}
Cordts, M., Omran, M., Ramos, S., Scharw{\"a}chter, T., Enzweiler, M.,
  Benenson, R., Franke, U., Roth, S., and Schiele, B., ``The cityscapes
  dataset,'' in [{\em CVPR Workshop on the Future of Datasets in
  Vision}{\nolinebreak\hspace{0.1em}]},   {\bf 2} (2015).

\bibitem{ros2016synthia}
Ros, G., Sellart, L., Materzynska, J., Vazquez, D., and Lopez, A.~M., ``The
  synthia dataset: A large collection of synthetic images for semantic
  segmentation of urban scenes,'' in [{\em Proceedings of the IEEE conference
  on computer vision and pattern recognition}{\nolinebreak\hspace{0.1em}]},
  3234--3243 (2016).

\bibitem{kingslake1989history}
Kingslake, R.,  [{\em A history of the photographic
  lens}{\nolinebreak\hspace{0.1em}]}, Elsevier (1989).

\bibitem{tardif2006self}
Tardif, J.-P., Sturm, P., and Roy, S., ``Self-calibration of a general radially
  symmetric distortion model,'' in [{\em European conference on computer
  vision}{\nolinebreak\hspace{0.1em}]},   186--199, Springer (2006).

\bibitem{kannala2006generic}
Kannala, J. and Brandt, S.~S., ``A generic camera model and calibration method
  for conventional, wide-angle, and fish-eye lenses,'' {\em IEEE transactions
  on pattern analysis and machine intelligence}~{\bf 28}(8),  1335--1340
  (2006).

\bibitem{saez2018cnn}
S{\'a}ez, A., Bergasa, L.~M., Romeral, E., L{\'o}pez, E., Barea, R., and Sanz,
  R., ``Cnn-based fisheye image real-time semantic segmentation,'' in [{\em
  2018 IEEE Intelligent Vehicles Symposium (IV)}{\nolinebreak\hspace{0.1em}]},
   1039--1044, IEEE (2018).

\bibitem{shah2018airsim}
Shah, S., Dey, D., Lovett, C., and Kapoor, A., ``Airsim: High-fidelity visual
  and physical simulation for autonomous vehicles,'' in [{\em Field and service
  robotics}{\nolinebreak\hspace{0.1em}]},   621--635, Springer (2018).

\bibitem{bourke2009idome}
Bourke, P., ``idome: Immersive gaming with the unity3d game engine,'' {\em
  CGAT09 Computer Games, Multimedia and Allied Technology}~{\bf 9},  265--272
  (2009).

\bibitem{UnityTec68:online}
``Unity-technologies / ml-imagesynthesis.''
\newblock (Accessed on 06/22/2019).

\bibitem{badrinarayanan2017segnet}
Badrinarayanan, V., Kendall, A., and Cipolla, R., ``Segnet: A deep
  convolutional encoder-decoder architecture for image segmentation,'' {\em
  IEEE transactions on pattern analysis and machine intelligence}~{\bf 39}(12),
   2481--2495 (2017).

\bibitem{haris_iqbal_2018_2526396}
Iqbal, H., ``Harisiqbal88/plotneuralnet v1.0.0,'' (Dec. 2018).

\bibitem{DBLP:journals/corr/ChenPSA17}
Chen, L., Papandreou, G., Schroff, F., and Adam, H., ``Rethinking atrous
  convolution for semantic image segmentation,'' {\em CoRR}~{\bf
  abs/1706.05587} (2017).

\bibitem{he2016identity}
He, K., Zhang, X., Ren, S., and Sun, J., ``Identity mappings in deep residual
  networks,'' in [{\em European conference on computer
  vision}{\nolinebreak\hspace{0.1em}]},   630--645, Springer (2016).

\bibitem{yu2015multi}
Yu, F. and Koltun, V., ``Multi-scale context aggregation by dilated
  convolutions,'' {\em arXiv preprint arXiv:1511.07122}  (2015).

\bibitem{chen2017deeplab}
Chen, L.-C., Papandreou, G., Kokkinos, I., Murphy, K., and Yuille, A.~L.,
  ``Deeplab: Semantic image segmentation with deep convolutional nets, atrous
  convolution, and fully connected crfs,'' {\em IEEE transactions on pattern
  analysis and machine intelligence}~{\bf 40}(4),  834--848 (2017).

\bibitem{ILSVRC15}
Russakovsky, O., Deng, J., Su, H., Krause, J., Satheesh, S., Ma, S., Huang, Z.,
  Karpathy, A., Khosla, A., Bernstein, M., Berg, A.~C., and Fei-Fei, L.,
  ``{ImageNet Large Scale Visual Recognition Challenge},'' {\em International
  Journal of Computer Vision (IJCV)}~{\bf 115}(3),  211--252 (2015).

\bibitem{glorot2010understanding}
Glorot, X. and Bengio, Y., ``Understanding the difficulty of training deep
  feedforward neural networks,'' in [{\em Proceedings of the thirteenth
  international conference on artificial intelligence and
  statistics}{\nolinebreak\hspace{0.1em}]},   249--256 (2010).

\bibitem{loshchilov2016sgdr}
Loshchilov, I. and Hutter, F., ``Sgdr: Stochastic gradient descent with warm
  restarts,'' {\em arXiv preprint arXiv:1608.03983}  (2016).

\bibitem{takahashi2018data}
Takahashi, R., Matsubara, T., and Uehara, K., ``Data augmentation using random
  image cropping and patching for deep cnns,'' {\em arXiv preprint
  arXiv:1811.09030}  (2018).

\bibitem{krizhevsky2012imagenet}
Krizhevsky, A., Sutskever, I., and Hinton, G.~E., ``Imagenet classification
  with deep convolutional neural networks,'' in [{\em Advances in neural
  information processing systems}{\nolinebreak\hspace{0.1em}]},   1097--1105
  (2012).

\bibitem{eichenseer2016data}
Eichenseer, A. and Kaup, A., ``A data set providing synthetic and real-world
  fisheye video sequences,'' in [{\em 2016 IEEE International Conference on
  Acoustics, Speech and Signal Processing
  (ICASSP)}{\nolinebreak\hspace{0.1em}]},   1541--1545, IEEE (2016).

\end{thebibliography}



\end{document}